\pdfoutput=1

\documentclass[11pt,table,usenames,dvipsnames]{article}

\usepackage{acl}

\usepackage{times}
\usepackage{latexsym}
\usepackage{csquotes}
\usepackage{tabularx}
\usepackage{color}
\usepackage{xcolor}
\usepackage{multirow}
\usepackage{caption}
\usepackage{graphicx}
\usepackage{hyperref}
\usepackage{soul}

\usepackage[T1]{fontenc}

\usepackage[utf8]{inputenc}

\usepackage{microtype}

\title{\textsc{AbLit}: A Resource for Analyzing and Generating\\ Abridged Versions of English Literature}

\author{Melissa Roemmele \quad Kyle Shaffer \quad Katrina Olsen \quad Yiyi Wang \quad Steve DeNeefe\\
        Language Weaver (RWS Group)\\
        \texttt{\string{mroemmele,kshaffer,kolsen,yiyiwang,sdeneefe\string}@rws.com}}

\begin{document}
\maketitle
\begin{abstract}
Creating an abridged version of a text involves shortening it while maintaining its linguistic qualities. In this paper, we examine this task from an NLP perspective for the first time. We present a new resource, \textsc{AbLit}, which is derived from abridged versions of English literature books. The dataset captures passage-level alignments between the original and abridged texts. We characterize the linguistic relations of these alignments, and create automated models to predict these relations as well as to generate abridgements for new texts. Our findings establish abridgement as a challenging task, motivating future resources and research. The dataset is available at \href{https://github.com/roemmele/AbLit}{github.com/roemmele/AbLit}.
\end{abstract}

\section{Introduction}


An abridgement is a shortened form of a text that maintains the linguistic qualities of that text\footnote{The term ``linguistic qualities'' is broad, which reflects other definitions of abridgement. For instance, the \href{https://en.wikipedia.org/wiki/Abridgement}{Wikipedia entry} for ``abridgement'' specifies that it ``maintains the unity of the source'', but these dimensions of unity are tacitly defined.}. It is intended to make the original text faster and easier to read. In this paper, we propose abridgement as an NLP problem and describe its connection to existing inference and generation tasks. We present a novel dataset for this task, focused on \underline{ab}ridged versions of English \underline{lit}erature books, which we refer to as the \textsc{AbLit} dataset. We demonstrate the characteristics of \textsc{AbLit} in terms of the relations between original and abridged texts as well as the challenges of automatically modeling these relations. The dataset and all associated code, including a Python package for easily interfacing with the data, are available at: \href{https://github.com/roemmele/AbLit}{github.com/roemmele/AbLit}. 

\section{The task of abridgement}\label{defining}

\subsection{Definition}\label{relation_to_tasks}

We define abridgement as the task of making a text easier to understand while preserving its linguistic qualities. As such, abridgement intersects with tasks that fuse natural language inference (NLI) and natural language generation (NLG), in particular summarization and simplification.


Summarization condenses the main content of a text into a shorter version in order to facilitate high-level comprehension of the content. Existing research has used the categories of \textit{extractive} and \textit{abstractive} to describe summaries. In the former, the summary `extracts' sequences from the text, whereas in the latter the summary `abstracts' out the meaning of the text and rewrites it. The degree of abstractiveness of a summary is indicated by how much novel text it contains that is not directly in the original text. Like a summary, an abridgement is shorter than its original text, but it preserves more of its language and can be seen as an alternative version rather than a meta-description. According to how summaries are characterized, abridgements are highly extractive, even if some abstraction is needed to connect the extracted components. 

Some work has examined summarization of narratives, including literary text \cite{kazantseva-szpakowicz-2010-summarizing,mihalcea-ceylan-2007-explorations,zhang2019}. Of particular relevance to our work are datasets released by \citet{chaudhury2019}, \citet{kryscinski2021booksum}, and \citet{ladhak-etal-2020-exploring}, all of which consist of summaries of fiction books. These summaries are significantly different from abridgements in that they are highly abstractive; they convey the book's narrative without preserving much of the text itself. \citeauthor{kryscinski2021booksum} provides summaries at different levels of granularity (book, chapter, and paragraph). Their analysis demonstrates that even the finer-grained summaries at the paragraph level are quite abstractive. 




The task of simplification also aims to make a text easier to understand, but without significantly distilling its content. Simplification is often treated as a sentence-level task \cite{sun-etal-2021-document}. Abridgement can be viewed as simplification on a document level. It seeks to balance the goal of increasing readability with preserving as much of the original text as possible. Research on simplification has been constrained by a lack of high-quality publicly available datasets. Existing datasets have been derived from sources like Wikipedia \cite[e.g.][]{coster-kauchak-2011-simple} and news articles \cite{xu-etal-2015-problems}, but none have focused on literary text. 

\subsection{Practical application}\label{applications}

There are few authors who perform abridgement, and thus relatively few abridged versions of books \cite{minshull2001}. Authors have described it as challenging and time-consuming to discern what to modify without compromising the original author's agency \cite{lauber1998,sussman1988}. However, as touted by these authors, abridgement makes books more accessible to a larger audience, especially when delivering the content through non-text modes like audio \cite{lavin2014}. Given this, automating the abridgement process could vastly expand the number of abridged versions of books and thus increase their readership. Automation does not preclude the involvement of human authors; for example, human translators use machine translation to increase their productivity \cite[e.g.][]{zhechev2014analysing}, and the same paradigm could apply to abridgement.

\section{Creating an abridgement dataset}\label{creating}

The \textsc{AbLit} dataset is derived from 10 classic English literature books, listed in \ref{book_sizes}. These books are in the public domain and available through Project Gutenberg\footnote{\href{https://www.gutenberg.org}{gutenberg.org}}. A single author, Emma Laybourn, wrote abridged versions of these books that are also freely available\footnote{\href{http://www.englishliteratureebooks.com/classicnovelsabridged.html}{englishliteratureebooks.com}}. The author explains:

\blockquote{\small{``This is a collection of famous novels which have been shortened and slightly simplified for the general reader. These are not summaries; each is half to two-thirds of the original length. I've selected works that people often find daunting because of their density or complexity: the aim is to make them easier to read, while keeping the style intact.''}}


Informed by this, we designed \textsc{AbLit} to capture the alignment between passages in a text and its abridged version. We specify that an abridged and original passage are aligned if the content of the abridged passage is fully derived from the original.

After obtaining the original and abridged books from their respective sites, we detected chapter headings to split the books into chapters (see \ref{appendix_chapter_boundaries} for details). We paired the original and abridged version of each chapter according to these headings. Obviously, the two versions already form a broad alignment unit, but our goal was to examine finer levels of alignment. We chose to use sentences as the minimal alignment units, since they are intuitive units of expression in text and can be detected automatically\footnote{We used \href{https://www.nltk.org/api/nltk.tokenize.html}{nltk.org} for all sentence segmentation and word tokenization. For analyses pertaining to words, words are lowercased without any other normalization (e.g. lemmatization).}. \textsc{AbLit} annotates sentence boundaries by indexing their position in the text, which enables all whitespace characters (most importantly, line breaks marking paragraphs) to be preserved.

\subsection{Automated alignments}\label{automated_alignments}


We pursued an automated approach to establish initial alignments between the original and abridged sentences for each chapter. It follows the same dynamic programming scheme used to create the Wikipedia Simplification dataset \cite{coster-kauchak-2011-simple}.  We refer to a group of adjacent sentences in a text as a \textit{span}. We define the length of a span by the number of sentences it contains. Each span $o$ of length $o_n$ in the original version of a chapter is paired with a span $a$ of length $a_m$ in the abridged version. The value of $a_m$ can be zero, allowing for the possibility that an original sentence is aligned with an empty string. 


For each pair of $o$ and $a$, we use a similarity metric $sim(o, a)$ to score the likelihood that they are aligned. This scoring function also considers the length of the spans in order to optimize for selecting the narrowest alignment between the original and abridged text. For instance, if a one-to-one alignment exists such that the meaning of a single sentence in the abridgement is fully encapsulated by a single original sentence, these sentences should form an exclusive alignment. To promote this, we adjust $sim(o,a)$ by a penalty factor $pn$ applied to the size of the pair, where ${size = \max(o_n, a_m)}$. Ultimately, the alignment score for a given span pair ($o$, $a$) is ${\max(0, sim(o, a) - ((size - 1) * pn))}$. Thus, more similar pairs obtain higher scores, but the scores are increasingly penalized as their size increases. At each sentence position in the original and abridged chapters, we score spans of all lengths $[1, o_n]$ and $[0, a_m]$, then select the one that obtains the highest score when its value is combined with the accumulated score of the aligned spans prior to that position. Once all span pairs are scored, we follow the backtrace from the highest-scoring span in the final sentence position to retrieve the optimal pairs for the chapter. We refer to each resulting span pair $(o, a)$ in this list as an alignment \textit{row}.



\subsection{Assessment of automated alignments}\label{assess_auto_alignments}

We applied this automated alignment approach to the first chapter in each of the ten books in \textsc{AbLit}, which we designated as an \textit{assessment} set for investigating the quality of the output rows. We instantiated $sim(o,a)$ as the ROUGE-1 (unigram) precision score\footnote{Using \href{https://github.com/Diego999/py-rouge}{github.com/Diego999/py-rouge}} between the spans, where $a$ is treated as the hypothesis and $o$ is treated as the reference. Here we refer to this score as $R$-$1_p$. It effectively counts the proportion of words in $a$ that also appear in $o$. We considered values of $o_n$ in $[1, 6]$ and $a_m$ in $[0, 6]$ and selected $o_n=3$ and $a_m=5$ based on our perceived quality of a sample of output rows. We similarly optimized $pn$ values in $[0, 0.25]$ and selected $pn$ = 0.175. Smaller values of $pn$ yielded rows that were not minimally sized (i.e. they needed to be further split into multiple rows), whereas larger values tended to wrongly exclude sentences from rows. 

The output consisted of 1,126 rows, which were then reviewed and corrected by five human validators recruited from our internal team. Validators judged a row as correct if the meaning expressed by the abridged span was also expressed in the original span, consistent with how alignment is defined above. \ref{validation_interface} gives more detail about this task. We found that inter-rater agreement was very high (Fleiss' $\kappa$ = 0.984) and the few disagreements were easily resolved through discussion. The validators reported spending 10-15 minutes on each chapter. 

After establishing these gold rows for the assessment set, we evaluated the initial automated rows with reference to the gold rows. To score this, we assigned binary labels to each pair of original and abridged sentences, where pairs that were part of the same row were given a positive class label and all other pairs were given a negative class label. Given these labels for the rows automatically produced with the $R$-$1_p$ method compared against the labels for the gold rows, the F1 score of the automated rows was 0.967. We also evaluated other methods for computing $sim(o, a)$,  but none outperformed $R$-$1_p$. See \ref{appendix_automated_alignment} for the description of these alternative methods and their results.


\begin{table*}[ht!]
\centering
\begin{tabularx}{\textwidth}{X p{0.41\textwidth}}
\textbf{Original Span} & \textbf{Abridged Span}\\
\hline

\small $[$\textcolor{CornflowerBlue}{\textbf{The letter}} was not unproductive.$]$ $[$It \textcolor{CornflowerBlue}{\textbf{re-established peace and kindness.}}$]$ & \small $[$\textcolor{CornflowerBlue}{\textbf{The letter re-established peace and kindness.}}$]$\\
\hline

\small $[$\textcolor{CornflowerBlue}{\textbf{Mr. Guppy sitting on the window-sill,}} nodding his head and balancing all these possibilities in his mind, continues \textcolor{CornflowerBlue}{\textbf{\underline{thoughtfully}}} to tap \textcolor{CornflowerBlue}{\textbf{it}}, and clasp it, and measure it with his hand \textcolor{CornflowerBlue}{\textbf{until he hastily draws his hand away.}}$]$ & \small $[$\textcolor{CornflowerBlue}{\textbf{Mr. Guppy sitting on the window-sill,}} \textcolor{ForestGreen}{\textbf{taps}} \textcolor{CornflowerBlue}{\textbf{it}} \textcolor{CornflowerBlue}{\textbf{\underline{thoughtfully}}}\textcolor{CornflowerBlue}{\textbf{, until he hastily draws his hand away.}}$]$\\
\hline
\small $[$\textcolor{CornflowerBlue}{\textbf{At last the gossips thought they had found the key to her conduct, and her uncle was sure of it}}; and what is more, \textcolor{CornflowerBlue}{\textbf{the discovery}} showed \textcolor{CornflowerBlue}{\textbf{\underline{his niece}}} to him in quite a new light, and he changed \textcolor{CornflowerBlue}{\textbf{his whole}} deportment \textcolor{CornflowerBlue}{\textbf{to}} her accordingly\textcolor{CornflowerBlue}{\textbf{.}}$]$ & \small $[$\textcolor{CornflowerBlue}{\textbf{At last the gossips thought they had found the key to her conduct, and her uncle was sure of it}} \textcolor{ForestGreen}{\textbf{.}}$]$ $[$\textcolor{CornflowerBlue}{\textbf{The discovery}} \textcolor{ForestGreen}{\textbf{altered}} \textcolor{CornflowerBlue}{\textbf{his whole}} \textcolor{ForestGreen}{\textbf{behaviour}} \textcolor{CornflowerBlue}{\textbf{to}} \textcolor{CornflowerBlue}{\textbf{\underline{his niece}.}}$]$\\
\hline
\small $[$\textcolor{CornflowerBlue}{\textbf{They trooped}} down into the hall and into the carriage, Lady Pomona leading the way\textcolor{CornflowerBlue}{\textbf{\underline{.}}}$]$ $[$\textcolor{CornflowerBlue}{\textbf{Georgiana}} stalked \textcolor{CornflowerBlue}{\textbf{along}}, passing \textcolor{CornflowerBlue}{\textbf{her father at the front door without condescending to look at him.}}$]$ & \small $[$\textcolor{CornflowerBlue}{\textbf{They trooped}} \textcolor{ForestGreen}{\textbf{downstairs}}, \textcolor{CornflowerBlue}{\textbf{Georgiana}} \textcolor{ForestGreen}{\textbf{stalking}} \textcolor{CornflowerBlue}{\textbf{along}}\textcolor{CornflowerBlue}{\textbf{\underline{.}}}$]$ $[$\textcolor{ForestGreen}{\textbf{She passed}} \textcolor{CornflowerBlue}{\textbf{her father at the front door without condescending to look at him.}}$]$\\
\hline
\end{tabularx}
\caption{Examples of alignment rows. Sentence boundaries are denoted by brackets ($[ ]$). We highlight preserved words in \textcolor{CornflowerBlue}{\textbf{blue}} and \underline{underline} the reordered ones. Added words are in \textcolor{ForestGreen}{\textbf{green}}.}
\label{examples}
\end{table*}

\subsection{Full dataset}\label{full_dataset}

\paragraph{Partial validation:} The time spent validating this assessment set indicated it would require significant resources to review the rows for all 868 chapters across the 10 books. Meanwhile, our evaluation revealed that we can expect most automatically aligned rows to be correct. Thus, we considered how to focus effort on correcting the small percentage of erroneously aligned rows. We manually reviewed these rows in the assessment set and found that their $R$-$1_p$ scores were often lower than those of the correct rows. Moreover, this tended to affect two types of rows: those with two or more sentences in the abridged span, or those adjacent to another row with an empty abridged span (i.e. $a_m$ = 0). We did an experiment where a human validator reviewed only the assessment rows with scores < 0.9 that qualified as one of the two above cases. Selectively applying corrections to just these rows boosted the F1 score of the assessment set from 0.967 to 0.99. We thus decided to apply this strategy of partially validating automated rows to create the training set for \textsc{AbLit}.

\paragraph{Final sets:} To construct the rest of \textsc{AbLit}, we ran the automated alignment procedure on all other chapters, and then applied the above partial validation strategy. Because we previously confirmed high inter-rater agreement, a single validator reviewed each chapter. Generalizing from the assessment set, we estimate that 99\% of these rows are correct. To ensure an absolute gold standard for evaluation, we set aside five chapters in each of the ten books and fully validated their rows. We designated this as the test set, and repurposed the assessment set to be a development set that we used accordingly in our experiments. All other chapters were assigned to the training set. Ultimately, \textsc{AbLit} consists of 808, 10, and 50 chapters in the training, development, and test sets, respectively. Table \ref{examples} shows some examples of rows in \textsc{AbLit}.

\section{Characterizing abridgements}\label{characterizing}

\subsection{Overview}

\begin{table}[h!]
\centering
\begin{tabularx}{\linewidth}{ p{1.15cm} | p{1.4cm} | p{0.95cm} | X }
& \multicolumn{1}{c}{\textbf{Train}} & \multicolumn{1}{c}{\textbf{Dev}} & \textbf{Test \scriptsize{(Chpt Mean)}}\\
\hline
\rowcolor{gray!20}Chpts & 808 & 10 & 50\\
\hline
Rows & 115,161 & 1,073 & 9,765 \footnotesize{(195)}\\
\hline
\rowcolor{gray!20}$O_{pars}$ & 37,227 & 313 & 3,125 \footnotesize{(62)}\\ 
\rowcolor{gray!20}$A_{pars}$ & 37,265 & 321 & 3,032 \footnotesize{(61)}\\
\hline
$O_{sents}$ & 122,219 & 1,143 & 10,431 \footnotesize{(209)}\\
$A_{sents}$ & 98,395 & 924 & 8,346 \footnotesize{(167)}\\
$\%A_{sents}$ & 80.5 & 80.8 &  80.0\\
\hline
\rowcolor{gray!20}$O_{wrds}$ & 2,727,571 & 29,908 & 231,878 \footnotesize{(4,638)} \\
\rowcolor{gray!20}$A_{wrds}$ & 1,718,919 & 17,630 & 143,908 \footnotesize{(2,878)} \\
\rowcolor{gray!20}$\%A_{wrds}$ & 63.0 & 58.9 & 62.1 \\
\hline
\end{tabularx}
\caption{Number of chapters (Chpts), alignment rows ($Rows$), paragraphs ($pars$), sentences ($sents$), and words ($wrds$) across all original ($O$) and abridged ($A$) books. The per-chapter means appear for the test set.}
\label{size_stats}
\end{table}

\begin{table}[h!]
\centering
\rowcolors{1}{white}{gray!20}
\begin{tabular}{c c | c c c}
\textbf{$O_{sents}$} & \textbf{$A_{sents}$} & \textbf{Train} & \textbf{Dev} & \textbf{Test}\\
\hline
1 & 1 & 75.8 & 74.7 & 75.7\\
1 & 0 & 17.4 & 17.3 & 17.3\\
2+ & 1 & 4.3 & 4.8 & 4.6\\
1 & 2+ & 2.1 & 3.2 & 1.9\\
2+ & 2+ & 0.3 & 0.0 & 0.5\\
\hline
\end{tabular}
\caption{Distribution of row sizes by number of sentences ($sents$) in original ($O$) and abridged ($A$) spans}
\label{row_size_stats}
\end{table}

Table \ref{size_stats} lists the size of \textsc{AbLit} in terms of rows, paragraphs, sentences, and words (see \ref{book_sizes} for these numbers compared by book). Here we call attention to the numbers for the fully-validated test set, but the numbers for the training set closely correspond. The development set slightly varies from the training and test set for a few statistics, likely because it is smaller. Judging by the test set, the abridged chapters have almost the same number of paragraphs as the original, but they have 80\% of the number of sentences ($\%A_{sents}$) and $\approx$62\% of the number of words ($\%A_{wrds}$). 

Table \ref{row_size_stats} pertains to the size of the original and abridged spans in each row, where size is the number of sentences in each span. The table shows the relative percentage of rows of each size. The majority of test rows ($\approx$76\%) contain a one-to-one alignment between an original and abridged sentence (i.e. $O_{sents}$ = 1, $A_{sents}$ = 1). Meanwhile, $\approx$17\% contain an original sentence with an empty abridged span ($O_{sents}$ = 1, $A_{sents}$ = 0). A minority of rows ($\approx$5\%) have a many-to-one alignment ($O_{sents}$ = 2+, $A_{sents}$ = 1) and a smaller minority ($\approx$2\%) have a one-to-many alignment ($O_{sents}$ = 1, $A_{sents}$ = 2+). Many-to-many alignments ($O_{sents}$ = 2+, $A_{sents}$ = 2+) are more rare (0.5\%).

\subsection{Lexical similarity}\label{lexical_similarity}

\begin{table}[h!]
\centering
\rowcolors{1}{white}{gray!20}
\begin{tabular}{c | c c c}
\textbf{Score Bin} & \textbf{Train} & \textbf{Dev} & \textbf{Test}\\
\hline
0.0 & 17.5 & 17.6 & 17.4\\
(0.0, 0.25] & 0.1 & 0.2 & 0.1\\
(0.25, 0.5] & 0.5 & 0.9 & 0.6\\
(0.5, 0.75] & 2.6 & 4.6 & 2.9\\
(0.75, 1.0) & 23.9 & 31.5 & 24.0\\
1.0 & 55.5 & 45.2 & 55.0\\
\hline
\end{tabular}
\caption{Binned distribution of $R$-$1_p$ scores for rows}\label{lexical_similarity_table}
\label{row_rouge_dist}
\end{table}

As demonstrated by the success of the $R$-$1_p$ metric for creating alignment rows (Section \ref{assess_auto_alignments}), an original and abridged span typically align if most of the words in the abridged are contained in the original. Table \ref{row_rouge_dist} shows the binned distribution of the $R$-$1_p$ scores for the rows. Rows with an exact score of 0.0 ($\approx$17\% of rows in the test set) consist almost exclusively of original spans aligned to empty spans, which is why this number is comparable to the second line of Table \ref{row_size_stats}. Many rows have perfect scores of exactly 1.0 (55\%), signifying that their abridged span is just an extraction of some or all of the original span. The abridged spans where this is not the case (i.e. they contain some words not in the original) still copy much of the original: 24\% of test rows have a $R$-$1_p$ score above 0.75 and below 1.0, while only a small minority ($\approx$4\%, the sum of lines 2-4 in the table) have a score above 0.0 and below 0.75.

\subsection{Lexical operations}\label{lexical_operations}

\begin{table}[h!]
\begin{tabularx}{\linewidth}{ p{1.9cm}  | X   | X  | X  }
& \textbf{Train} & \textbf{Dev} & \textbf{Test}\\
\hline
\rowcolor{gray!20}$O_{rmv}$ \scriptsize{($O_{prsv}$)} & 40.9 \scriptsize(59.1) & 45.9 \scriptsize(54.1) & 41.9 \scriptsize(58.1)\\
$A_{add}$ \scriptsize{($A_{prsv}$)} & 6.3 \scriptsize(93.7) & 8.3 \scriptsize(91.7) & 6.4 \scriptsize(93.6)\\
\hline

\rowcolor{white}\multicolumn{4}{c}{}\\
\hline
\rowcolor{gray!20}$Rows_{rmv}$ & 71.1 & 80.3 & 73.2\\
$Rows_{prsv}$ & 82.5 & 82.7 & 82.6\\
\rowcolor{gray!20}$Rows_{add}$ & 37.4 & 48.8 & 39.4\\
$Rows_{reord}$ & 11.8 & 16.5 & 11.7\\
\hline
\end{tabularx}
\caption{\textit{Top:} the \% of removed and added words relative to all original and abridged words, respectively. \textit{Bottom:} the \% of rows with each lexical operation.}
\label{operation_stats}
\end{table}

For each row, we enumerate the common and divergent items between the words $o_{wrds}$ in the original span and the words $a_{wrds}$ in the abridged span. The words that appear in $o_{wrds}$ but not $a_{wrds}$ are removed words, i.e. $o_{rmv} = {|o_{wrds} - a_{wrds}|}$. All other original words are preserved in the abridgement, i.e. ${o_{prsv} = |o_{wrds} - o_{rmv}|}$. Accumulating these counts across all original spans ${o \in O}$, the top section of Table \ref{operation_stats} indicates the percentages of removed and preserved words among all original words. In the test set, $\approx$42\% of words are removed, and thus $\approx$58\% are preserved. Next, we count the added words in the abridgement, which are those that appear in $a_{wrds}$ and not $o_{wrds}$, i.e. ${a_{add} = |a_{wrds} - o_{wrds}|}$. All other abridged words are preserved from the original, i.e. ${a_{prsv} = |a_{wrds} - a_{add}|}$. Accumulating these counts across all abridged spans ${a \in A}$, Table \ref{operation_stats} shows that only $\approx$6\% of abridged words in the test set are additions, and thus $\approx$94\% are preservations.

We also report the number of rows where these removal, preservation, and addition operations occur at least once. For instance, if $o_{rmv} > 0$ for the original span in a given row, we count that row as part of $Rows_{rmv}$. The bottom section of Table \ref{operation_stats} shows the percentage of rows with each operation among all rows in the dataset. In $\approx$73\% of the test rows, the abridged span removes at least one word from the original. In $\approx$83\% of rows, the abridged span preserves at least one word from the original. In $\approx$39\% of rows, the abridged span adds at least one word not in the original. We considered the possibility that preserved words could be reordered in the abridgement. To capture this, we find the longest contiguous sequences of preserved words (i.e. ``slices'') in the abridged spans. A row is included in $Rows_{reord}$ if at least two abridged slices appear in a different order compared to the original span. This reordering occurs in $\approx$12\% of rows. 

It is clear from this analysis that the abridgements are quite loyal to the original versions, but they still remove a significant degree of text and introduce some new text. The examples in Table \ref{examples} highlight these operations. We can qualitatively interpret from the examples that some added words in the abridged span are substitutions for removed original words (e.g. ``tap'' > ``taps'' in the second example, ``changed'' > ``altered'' in the third example). See \ref{nli} for additional discussion about how some of these relations pertain to common NLI tasks.

\subsection{Lexical categories}\label{lexical_categories}

\begin{table}[h!]
\centering
\rowcolors{1}{white}{gray!20}
\begin{tabularx}{\linewidth}{p{1.5cm} X p{1.2cm} | X p{1.2cm}}
\textbf{Category} & \textbf{\%$O$} & \textbf{\%$O_{rmv}$} & \textbf{\%$A$} & \textbf{\%$A_{add}$}\\
\hline
Function & 58.2 & 57.9 & 58.1 & 53.9\\
Content & 41.8 & 42.1 & 41.9 & 46.1\\
\hline
\end{tabularx}
\caption{Test set distribution of lexical categories for removed words $O_{rmv}$ compared with all original words $O$, and added words $A_{add}$ compared with all abridged words $A$}
\label{lexical_categories_table}
\end{table}

We examined if certain types of words are more often affected by removal or addition operations. Table \ref{lexical_categories_table} contains a broad analysis of this for the test set. As shown, $\approx$58\% of original words $O$ are function words (those with part-of-speech tags of punctuation, pronouns, adpositions, determiners, etc.), while $\approx$42\% are content words (nouns, verbs, adjectives, and adverbs). The category distribution of removed words $O_{rmv}$ closely matches the $O$ distribution, suggesting that both function and content words are removed at the same rate. The abridged words $A$ have the same proportion of function and content words as $O$ (again, $\approx$58\% and $\approx$42\%). In comparison, $\approx$54\% of additions $A_{add}$ are function words, while $\approx$46\% are content words. The gap between $\approx$42\% and $\approx$46\% indicates that content words are added at a slightly higher rate than the overall frequency in content words in $A$ (and equivalently, function words are added at a lower rate). But there are few additions overall, so the abridgements retain the same word type distribution as the original texts. \ref{extended_lexical_analysis} shows this same analysis for each specific part-of-speech tag among these types.

\section{Predicting what to abridge}\label{predicting}

\citet{garbacea-etal-2021-explainable} points out that a key (and often neglected) preliminary step in simplification is to distinguish text that could benefit from being simplified versus text that is already sufficiently simple. This is also an important consideration for abridgement, since it seeks to only modify text in places where it improves readability. Accordingly, we examine whether we can automatically predict the text in the original that should be removed when producing the abridgement. As explained in Section \ref{characterizing}, a removed word could mean the author replaced it with a different word(s) in the abridgement, or simply excluded any representation of its meaning. However, both cases indicate some change is applied to that word. 

We model this through a binary sequence labeling task. Given a passage with original tokens $o_{toks}$ and corresponding abridged tokens $a_{toks}$, we assign each token $t$ in $o_{toks}$ the label of \texttt{preserved} ($l$=0) if it is also in $a_{toks}$, and otherwise the label of \texttt{removed} ($l$=1) if it is not in $a_{toks}$. Thus the task is to predict the label sequence $[l_1, l_2, . . . l_n]$ from the token sequence $[t_1, t_2, . . . t_n]$.

\subsection{Model inputs}\label{model_inputs}

We can derive a token-label sequence from each alignment row, by which each original span corresponds to a single input instance. However, the size of these spans varies across rows. To produce models that handle texts where these span boundaries are not known in advance, we consider consistent-length passages whose boundaries can be automatically inferred. Thus the \textsc{AbLit} interface can provide pairs where a fixed-length passage from the original chapter (i.e. a sentence, paragraph, or multi-paragraph chunk) is aligned to its specific corresponding abridged version. 

We enable this by finding the respective positions of the longest common word sequences between the original and abridged spans. Each of these overlapping subsequences is represented as a slice of the original text with indices ($o_i$, $o_j$) mapped to a slice of the abridged text ($a_i$, $a_j$). Then, given a passage in the original text with indices ($o_l$, $o_m$), we find all enclosed slices ($o_i$, $o_j$) where $o_i >= o_l$ and $o_j <= o_m$. For each slice we retrieve its corresponding abridged slice ($a_i$, $a_j$). Given the earliest text position $\min a_i$ and latest position $\max a_j$ among these abridged slices, the full abridgement for the passage at ($o_l$, $o_m$) is the text covered by the indices ($\min a_i$, $\max a_j$). As an example, consider the first line in Table \ref{examples}. If retrieving abridgements for sentence-length passages, the first sentence in the original span ``The letter was not unproductive.'' will yield ``The letter'' as the abridgement. The second original sentence ``It re-established peace and kindness'' will yield the abridgement ``re-established peace and kindness''. By varying the passage size, we can assess how much context beyond a single row is beneficial in modeling abridgements. See \ref{model_inputs_comment} for more details.


\subsection{Experiment}

\paragraph{Model:} To predict abridgement labels (\texttt{preserved}/\texttt{removed}), we use a \textsc{RoBERTa}-based sequence labeling model, which has been applied to several other NLI tasks \cite{liu2019roberta}. We divided chapters according to varying passage sizes and trained a separate model on the token-label sequences\footnote{ A ``token'' in this case is a sub-token unit defined by the \textsc{RoBERTa} tokenizer, rather than a whitespace-separated ``word'' pertaining to Section \ref{characterizing}.} associated with each passage size. The passages are either sentences (detected by NLTK), paragraphs (detected by line breaks), or multi-paragraph `chunks'. Each chunk consists of one or more paragraphs of $S$ sentences, such that paragraphs are combined into the same chunk when their total number of sentences does not exceed $S$. As an additional reference, we trained a model where each passage is an original span directly taken from a single alignment row. As explained in Section \ref{model_inputs}, these passages (termed Rows) vary in length. We did not train a model on the full chapters as inputs because the average length of these inputs (5,044 tokens) greatly exceeds the \textsc{RoBERTa} limit of 512. See \ref{prediction_model_details} for more details about the model.

\begin{table}[h!]
\centering
\rowcolors{1}{white}{gray!20}
\begin{tabularx}{\linewidth}{p{2.4cm} | X X X X}
\textbf{Passage} & \textbf{Toks} & \textbf{P} & \textbf{R} & \textbf{F1}\\
\hline
Rows & 26 & 0.692 & 0.442 & 0.532\\
Sentences & 24 & 0.677 & 0.453 & 0.535\\
Paragraphs & 81 & 0.686 & 0.460 & 0.546\\
Chunks ($S$=10) & 303 & 0.670 & 0.501 & 0.569\\
\hline

\rowcolor{yellow!20}All=\texttt{removed}& - & 0.415 & 1.000 & 0.583\\
\hline
\end{tabularx}
\caption{F1 scores of abridgement label prediction for test set with models trained on varying passage sizes. \textbf{Toks} is the mean number of tokens per passage type.}
\label{prediction_results}
\end{table}

\paragraph{Results:} Each model is evaluated on instances of the corresponding passage size in the test set. Table \ref{prediction_results} displays the results in terms of the precision (P), recall (R), and F1 score of predicting that a token should be removed. We compare these results with the baseline of labeling all tokens in the chapter as \texttt{removed} (final line). For chunks, we tuned different values of $S$ in $[5, 11]$ on the development set and observed the best F1 at $S$=10. The results show that the longest passage size (Chunks) yields the best predictions, suggesting the importance of chapter context beyond that given in a single row. The consistently higher precision over recall for all models indicates they correctly predict many preservation operations, but at the expense of missing many removal operations. Consequently, they overestimate the number of tokens that should be preserved. This results in an overall F1 that is lower than what occurs when all tokens are removed.


\section{Producing abridgements}\label{producing}

The above results show that anticipating what parts of a text should be changed when writing its abridged version is not trivial. The full task of producing an abridgement implicitly involves inferring these \texttt{preserved}/\texttt{removed} labels while additionally predicting the specific text that dictates these labels. We examine models that have been applied to tasks related to abridgement to establish benchmarks for this new task, with the intent that these benchmarks will inspire future work.

\subsection{Models}

We consider the following models to produce an abridged version of an original chapter:

\paragraph{Naive Baselines:} As a reference point for our evaluation metrics, we report the performance of very weak baselines. In particular, we copy the entire original text as the abridgement (\textsc{Copy}). Alternatively, we select $T$ percent of original tokens (\textsc{RandExtToks}) as the abridgement. 

\paragraph{Extractive Approaches:} The analysis in Section \ref{lexical_similarity} showed that abridgements preserve much of their original text, which motivates the use of extractive summarization methods. Using the best label prediction model from Section \ref{predicting}, we extract all original tokens labeled as \texttt{preserved} to form the abridgement (\textsc{ExtToks}). To reveal the maximum performance that can be obtained with this method, we also run it using the gold labels instead of predicted labels (\textsc{PerfectExtToks}). It is not conventional to use tokens as units of extraction, since it can compromise fluency within sentences. \textsc{ExtToks} and \textsc{PerfectExtToks} only serve as points of comparison for our evaluation metrics. The standard extractive approach uses sentences as extractive units. For this  (\textsc{ExtSents}), we form an abridgement by selecting a subset of sentences in the original chapter where at least $P$ percent of tokens are labeled as \texttt{preserved}.

\paragraph{Generation Models:} Extractive methods cannot introduce words into the abridgement that are not in the original, so for this we need to consider generation models. In particular, we examine two transformer-based sequence-to-sequence models that have been used for various generation tasks including summarization:  \textsc{T5-base} \cite{raffel2020exploring} (termed \textsc{TunedT5} here) and \textsc{BART-base} \cite{lewis-etal-2020-bart} (termed \textsc{TunedBART}). We fine-tuned both models on the \textsc{AbLit} training set, specifically on inputs consisting of chunks with 10 sentences, since this passage size yielded the best results in the Section \ref{predicting} experiment. To assess the impact of these models' observation of \textsc{AbLit}, we compare them with abridgements produced by prompting the non-finetuned \textsc{T5-base} to perform zero-shot summarization (\textsc{ZeroshotT5}). See \ref{generation_models_details} for more details about these models. For all models, we generated an abridgement for an original chapter by dividing the chapter into chunks, generating output for each chunk (with 5-beam decoding), then concatenating the outputs to form the complete abridgement.

\subsection{Evaluation metrics}



We evaluate the predicted abridgements through comparison with the human-authored reference abridgements. First, we measure the word-based similarity between the predicted abridgement $a_{pred}$ and reference abridgement $a_{ref}$ using ROUGE-L ($R$-$L$), a standard evaluation metric for summarization. We then assess how accurately $a_{pred}$ removed and preserved words from the original. A word from the original in $a_{pred}$ is considered correctly preserved if it also appears in $a_{ref}$. We report the F1 of this measure as $Prsv$. A word in the original but not in $a_{pred}$ is considered correctly removed if it is also absent from $a_{ref}$. We report the F1 of this measure as $Rmv$. Finally, we evaluate the accuracy of added words, where a word not in the original is considered correctly added to $a_{pred}$ if it is also in $a_{ref}$. We report the F1 of this measure as $Add$. See \ref{evaluation_metrics_detail} for formal definitions of these metrics. 

\definecolor{light-light-gray}{gray}{0.9}

\begin{table*}[h!]
\centering
\begin{tabularx}{\textwidth}{p{3.65cm} p{5.5cm} | X X X X X}
\textbf{Name} & \textbf{Description} & \textbf{Toks} & \textbf{$R$-$L$} & \textbf{$Prsv$} & \textbf{$Rmv$} & \textbf{$Add$}\\
\hline
\rowcolor{yellow!20}\textsc{Human} & \small{Reference ($a_{ref}$)} & 2,878 & - & - & - & -\\
\hline
\rowcolor{gray!20}\textsc{Copy} & \small{Duplicate original} & 4,638 & 0.739 & 0.753 & 0.000 & 0.000\\
\rowcolor{gray!20}\textsc{RandExtToks} \small{($T$=0.6)} &  \small{$T\%$ randomly selected original tokens} & 2,787 & 0.753 & 0.800 & 0.694 & 0.000\\
\hline
\textsc{ExtToks} & \small{Original tokens predicted as \texttt{preserved}} & 3,160 & 0.818 & 0.856 & 0.745 & 0.006\\
\rowcolor{yellow!20}\textsc{PerfectExtToks} & \small{Original tokens where gold label is \texttt{preserved}} (upper bound for \textsc{ExtToks})  & \textcolor{darkgray}{2,664} & 0.950 & 0.969 & 0.954 & 0.034\\
\textsc{ExtSents} \small{($P$=0.65)} & \small{Original sentences with $\geq$ $P\%$ tokens predicted as \texttt{preserved}} & 2,857 & 0.792 & 0.824 & 0.720 & 0.001\\
\hline
\rowcolor{gray!20}\textsc{TunedT5} & \small{Generate from finetuned T5} & 3,834 & 0.727 & 0.804 & 0.519 & 0.275\\
\rowcolor{gray!20}\textsc{TunedBART} & \small{Generate from finetuned BART} & 3,673 & 0.780 & 0.815 & 0.583 & 0.365\\
\rowcolor{gray!20}\textsc{ZeroshotT5} & \small{Generate from non-finetuned T5} & 1,157 & 0.416 & 0.484 & 0.627 & 0.019\\
\hline
\end{tabularx}
\caption{Scores of predicted abridgements on evaluation metrics. For all metrics, higher scores are better.}
\label{generation_results}
\end{table*}

\begin{table*}[h!]
\centering
\small
\begin{tabularx}{\textwidth}{>{\columncolor{gray!10}}p{5.9cm} | >{\columncolor{green!10}}p{3.3cm} | >{\columncolor{violet!10}}X}
\textbf{Original} & \textbf{Reference} & \textbf{\textsc{TunedBART}}\\
\hline


The windows were half open because of the heat, and the Venetian blinds covered the glass,--so that a gray grim light, reflected from the pavement below, threw all the shadows wrong, and combined with the green-tinged upper light to make even Margaret's own face, as she caught it in the mirrors, look ghastly and wan. & \small The windows were half open because of the heat, and Venetian blinds covered the glass, giving the light a green tinge that made her face in the mirrors look ghastly and wan. & \small The windows were half open because of the heat, and the Venetian blinds covered the glass - so that a grey grim light, reflected from the pavement below, threw all the shadows wrong, and made even Margaret's own face look ghastly and wan.\\
\hline
We must suppose little George Osborne has ridden from Knightsbridge towards Fulham, and will stop and make inquiries at that village regarding some friends whom we have left there. How is Mrs. Amelia after the storm of Waterloo? Is she living and thriving? What has come of Major Dobbin, whose cab was always hankering about her premises? & \small We must now make inquiries at Fulham about some friends whom we have left there. How is Mrs. Amelia? Is she living and thriving? What has become of Major Dobbin? & \small We must suppose little George Osborne has ridden towards Fulham, and will stop and make inquiries about some friends whom we have left there. How is Mrs. Amelia after the storm of Waterloo? Is she living and thriving? What has come of Major Dobbin, whose cab was always hankering about her premises?\\

\hline
\end{tabularx}
\caption{Abridgements predicted by \textsc{TunedBART} for excerpts of North and South and Vanity Fair}
\label{pred_abridgement_examples}
\end{table*}

\subsection{Results}\label{results}

Table \ref{generation_results} reports the length and metric scores of the abridgements produced by each model for the test set chapters. Where applicable, we selected the $T$ and $P$ parameters from tuning on the development set. The results again convey that abridgement is largely a text extraction task, though a challenging one. The low $R$-$L$ score of \textsc{ZeroShotT5} confirms that \textsc{AbLit} is different from the summarization datasets that \textsc{T5-base} is trained on. The high $R$-$L$ of \textsc{PerfectExtToks} validates that precisely identifying which words to remove goes far in producing the abridgement. The high $Prsv$ scores for all approaches that observe \textsc{AbLit} show they can all preserve the original text reasonably well. Analogous to the results in Section \ref{predicting}, the lower $Rmv$ scores indicate knowing which words to remove is harder, particularly for the generation models. The extractive methods have no opportunity to obtain an $Add$ score that is non-trivially above 0\footnote{It is possible for $Add$ to be slightly above 0 with the extractive approaches due to tokenization; see \ref{evaluation_metrics_detail}.}. The generation models do show a small benefit here in correctly adding some new words to the abridgement. The examples in Table \ref{pred_abridgement_examples} qualitatively represent the outcome for the \textsc{TunedBART} model. These abridgements remove some of the same original text as the reference and also add a few words consistent with the reference, but they still retain more of the original text than the reference. Other examples are shown in \ref{produced_abridgement_examples}.

\section{Conclusion}\label{conclusion}

In this paper, we introduced \textsc{AbLit}, a corpus of original and abridged versions of English literature. \textsc{AbLit} enables systematic analysis of the abridgement task, which has not yet been studied from an NLP perspective. Abridgement is related to other tasks like summarization, but has a stricter requirement to maintain loyalty to the original text. Our experiments motivate an opportunity to better balance this goal against that of improving readability. We also envision future resources that generalize this task to other texts beyond English literature.

\section{Limitations}

We present \textsc{AbLit} to introduce abridgement as an NLP task. However, the dataset is scoped to one small set of texts associated with a specific domain and author. 

There are significant practical reasons for this limited scope. In particular, most recently published books are not included in publicly accessible datasets due to copyright restrictions, and the same restrictions typically apply to any abridgements of these books. The books in \textsc{AbLit} are uniquely in the public domain due to expired copyrights, and the author chose to also provide her abridgements for free. For this reason, \textsc{AbLit} consists of British English literature from the 18th and 19th centuries. Some of the linguistic properties of these original books do not generalize to other types of English texts that would be useful to abridge. We do not yet know what aspects of abridgement are specific to this particular domain.

Moreover, as described in Section \ref{applications}, creating abridgements is a rare and highly skilled writing endeavor. The \textsc{AbLit} abridgements are written exclusively by one author. Without observing alternative abridgements for the same books by a different author, it is unclear what features are specific to the author's preferences. This conflation between task and author is a concern for many NLP datasets \cite{geva-etal-2019-modeling}. More generally, obtaining human writing expertise is a challenge shared by all language generation research as it becomes more ambitious \cite[e.g.][]{wu2021recursively}.

\section{Ethical Considerations}

As stated in the introduction, all data and code used in this work is freely available. The text included in the dataset is in the public domain. Additionally, we explicitly confirmed approval from the author of the abridged books to use them in our research. 

For the data validation task, the validators were employed within our institution and thus were compensated as part of their normal job role. Given that the dataset is derived directly from published books, it is possible that readers may be offended by some content in these books. The validators did not report any subjective experience of this.

With regard to our modeling approaches, large pretrained models like the ones we use here for generating abridgements have a well-known risk of producing harmful content \cite[e.g.][]{gehman-etal-2020-realtoxicityprompts}. For the generation models fine-tuned on \textsc{AbLit}, we did not subjectively observe any such text in the sample output we assessed. We judge that our controlled selection of training data reduces this risk, but does not eliminate it. Accordingly, future applications of abridgement can similarly consider careful data curation for mitigating this risk. 

\bibliography{anthology,custom}
\bibliographystyle{acl_natbib}

\appendix

\clearpage
\section{Appendix}
\label{sec:appendix}

\subsection{Detecting chapter boundaries}\label{appendix_chapter_boundaries}

There is a one-to-one relation between each chapter in an original book and each chapter in its corresponding abridged version. Both the original and abridged version of the books include headings separating chapters. We automatically detected these headings through a set of regular expressions (e.g. matching lines specifying a chapter number and name with the regular expression \mbox{``\texttt{\string^Chapter [0-9]+:*[a-zA-Z\textbackslash s]*\$}''}). However, there is variability in the format of the headings: some can span multiple lines, or specify a book and volume number in addition to the chapter identifier, or have numbers written in non-numerical form, for instance. The format also varies between the original and abridged version of the same book. Thus, we manually reviewed all detected chapter boundaries and fixed any erroneous or missed boundaries. Ultimately we ensure that each chapter in an original book is paired exactly with its abridged counterpart.

\subsection{Additional automated alignment results}\label{appendix_automated_alignment}

\begin{table*}[h!]
\centering
\begin{tabular}{p{4.1cm} c c c c}
\textbf{Similarity Metric} & \textbf{$pn$} & \textbf{P} & \textbf{R} & \textbf{F1}\\
\hline
\multicolumn{5}{l}{\cellcolor{gray!30}\textit{Vector cosine similarity}}\\
\textsc{BERT-base-uncased} & 0.21 & 0.963 & 0.952 & 0.957\\
\textsc{BERT-base-cased} & 0.22	& 0.948 & 0.934 & 0.940\\
\textsc{BERT-large-uncased}	& 0.21 & 0.934 & 0.919 & 0.926\\
\textsc{BERT-large-cased} & 0.21 & 0.944 & 0.935 & 0.939\\
\textsc{XLNet-base-cased} & 0.22 & 0.753 & 0.731 & 0.742\\
\textsc{XLNet-large-cased} & 0.21 & 0.583 & 0.564 & 0.573\\
\textsc{XLM-mlm-en} & 0.21 & 0.821 & 0.816 & 0.818\\
\textsc{RoBERTa-base} & 0.21 & 0.738 & 0.717 & 0.727\\
\textsc{RoBERTa-large} & 0.21 & 0.592 & 0.573 & 0.582\\
\hline
\multicolumn{5}{l}{\cellcolor{gray!30}\textit{Word overlap similarity}}\\
\textsc{$R$-$1_p$} & 0.175 & 0.964 & 0.969 & 0.967\\
\textsc{$R$-$2_p$} & 0.175 & 0.912 & 0.958 & 0.935\\
\hline
\rowcolor{yellow!8}\textsc{$R$-$1_p$} \newline+ partial human validation & 0.175 & 0.990 & 0.991 & 0.990\\
\hline
\end{tabular}
\caption{Extended results for accuracy of automated alignment methods}
\label{automated_alignment_results_full}
\end{table*}

Table \ref{automated_alignment_results_full} shows the results of all methods we assessed for computing similarity between original and abridged spans to create alignment rows, compared alongside the best method of unigram ROUGE precision ($R$-$1_p$) reported in Section \ref{assess_auto_alignments}. A clear drawback to using unigram overlap to measure similarity is that it does not account for differences in word order. However, taking this into account by using bigrams instead of unigrams to calculate ROUGE precision (i.e. $R$-$2_p$) reduced the F1 to 0.935, likely because it added more sparsity to the overlap units. In addition to the word-based ROUGE metric, we assessed vector-based similarity encoded by different configurations of pretrained language models: \textsc{BERT} \cite{devlin-etal-2019-bert}, \textsc{XLNet} \cite{yang2019xlnet}, \textsc{XLM} \cite{conneau2019cross}, and \textsc{RoBERTa} \cite{liu2019roberta}. We used the HuggingFace Transformers implementation of these models: \url{https://huggingface.co/docs/transformers/index}. For each model we report the best result among size penalty ($pn$) values in $[0, 0.25]$. As displayed, the vectors that obtained the best F1 came from \textsc{BERT} \cite{devlin-etal-2019-bert}, particularly \textsc{bert-base-uncased}, which consists of 110M parameters. See additional details about this model here: \url{https://huggingface.co/bert-base-uncased}. Ultimately, however, the result from \textsc{bert-base-uncased} was still outperformed by $R$-$1_p$. As reported in Section \ref{full_dataset}, the resulting rows were further improved by applying the described partial validation strategy (final line of table).

\subsection{Details about validation task}\label{validation_interface}

For each row, validators assessed whether the abridged span in the row was correctly aligned with the corresponding original span. As described in Section \ref{assess_auto_alignments}, a row is correct if the meaning of the abridged span can be derived from the original span. For a given row, if the abridged span expressed some meaning not contained in the original span, it either meant that some sentences(s) in the abridged chapter were incorrectly placed in that row, or some sentence(s) in the original chapter were incorrectly placed in a different row. In both cases, validators moved the wrongly placed sentence(s) to a row resulting in correctly aligned spans. We utilized Google Sheets as an interface for this task, which enabled validators to easily review and correct the rows. We produced a single spreadsheet per chapter, where each spreadsheet row corresponded to an alignment row. For the partial validation strategy, we designed a Google Apps Script (\url{https://developers.google.com/apps-script}) that visually highlighted spreadsheet rows qualifying for partial validation so that validators could specifically attend to those rows.

For the development (assessment) and test sets, there were a few cases where the validators edited the spans themselves in order to correct sentence segmentation errors (e.g. wrongly segmenting after honorifics like ``Mr.''). 

\subsection{Size of \textsc{AbLit} compared by book}\label{book_sizes}

Table \ref{book_level_size_stats} shows characteristics of the data for each book in terms of number of alignment rows, original words, and abridged words.

\begin{table*}[h!]
\rowcolors{1}{white}{gray!20}
\centering
\begin{tabularx}{\textwidth}{ p{1.95cm} | p{1.1cm}  p{1.35cm} X | p{0.75cm}  p{0.9cm}  X | p{0.9cm}  X  X}
& \multicolumn{3}{ c }{\textbf{Train}} & \multicolumn{3}{ c }{\textbf{Dev}} & \multicolumn{3}{ c }{\textbf{Test}}\\
\cline{2-10}
Book \newline\scriptsize{ (Orig Author)} & Rows \newline(Chpts) & $O_{wrds}$ & \%$A_{wrds}$ & Rows & $O_{wrds}$ & \%$A_{wrds}$ & Rows & $O_{wrds}$ & \%$A_{wrds}$\\
\hline
Bleak House \newline\scriptsize{(Charles Dickens)} & 17,948 (62) & 390,857 & 63.2 & 24 & 935 & 20.0 & 1,746 & 38,132 & 62.9\\
Can You Forgive Her? \newline\scriptsize{(Anthony Trollope)} & 16,494 (74) & 350,092 & 62.2 & 94 & 3,216 & 49.5 & 1,339 & 27,660 & 61.2\\
Daniel Deronda  \newline\scriptsize{(George Eliot)} & 12,735 (64) & 333,283 & 61.6 & 158 & 3,524 & 61.9 & 786 & 25,334 & 49.1\\
Mansfield Park  \newline\scriptsize{(Jane Austen)} & 5,744 (42) & 159,863 & 67.0 & 91 & 3,564 & 62.1 & 795 & 22,607 & 66.1\\
North and South  \newline\scriptsize{(Elizabeth Gaskell)} & 8,922 (46) & 193,355 & 67.9 & 184 & 4,907 & 68.5 & 1,169 & 23,159 & 70.0\\
Shirley  \newline\scriptsize{(Charlotte Bronte)} & 12,027 (31) & 235,888 & 63.2 & 253 & 5,987 & 57.4 & 1,031 & 23,369 & 60.4\\
The Way We Live Now \newline\scriptsize{(Anthony Trollope)} & 19,355 (94) & 392,554 & 60.3 & 166 & 4,345 & 53.7 & 1,122 & 23,238 & 60.7\\
Tristram Shandy  \newline\scriptsize{(Laurence Sterne)} & 4,805 (305) & 216,984 & 66.7 & 5 & 439 & 77.0 & 69 & 3,972 & 72.3\\
Vanity Fair  \newline\scriptsize{(W. M. Thackeray)} & 11,682 (62) & 334,783 & 59.8 & 18 & 717 & 60.9 & 738 & 23,609 & 57.4 \\
Wuthering Heights  \newline\scriptsize{(Emily Bronte)} & 5,449 (28) & 119,912 & 66.3 & 80 & 2,274 & 68.3 & 970 & 20,798 & 71.0\\
\hline
\rowcolor{gray!40}All & 115,161 (808) & 2,727,571 & 63.0 & 1,073 & 29,908 & 58.9 & 9,765 & 231,878 & 62.1\\
\hline
\end{tabularx}
\caption{Statistics for each book in the AbLit dataset, in terms of number of alignment rows, total original word ($O_{wrds}$), and proportional length of abridgement relative to original ($\%A_{wrds}$). The number of chapters in the training set for each book is shown; there is 1 chapter per book in the development set and 5 chapters per book in the test set.}
\label{book_level_size_stats}
\end{table*}


\subsection{NLI challenges in \textsc{AbLit}}\label{nli}

Table \ref{harder_examples} shows some examples of rows in \textsc{AbLit} where modeling the relation between the original and abridged span involves NLI challenges like abstractive paraphrasing, figurative language interpretation, commonsense reasoning, and narrative understanding.

\begin{table*}[h!]
\rowcolors{1}{white}{gray!20}
\centering
\begin{tabularx}{\textwidth}{X p{3.75cm} X}
\textbf{Original Span} & \textbf{Abridged Span} & \textbf{Type of Challenge}\\
\hline
Still there was not a word. & No one spoke. & Paraphrasing: abridgement has same meaning as original but no word overlap\\
\hline
But it is time to go home; my appetite tells me the hour. & But it is time to go home; I am hungry. & Interpretation of figurative language: abridgement replaces phrase ``appetite tells me the hour'' with more literal term ``hungry''\\
\hline
``Daniel, do you see that you are sitting on the bent pages of your book?'' & ``Daniel, you are sitting on the bent pages of your book.'' & Change in dialogue act: question in original is transformed into statement in abridgment\\
\hline
While she was at Matching, and before Mr. Palliser had returned from Monkshade, a letter reached her, by what means she had never learned. ``A letter has been placed within my writing-case,'' she said to her maid, quite openly. ``Who put it there?'' & While she was at Matching, a letter reached her, by what means she never learned, although she suspected her maid of placing it inside her writing-case. & Dialogue interpretation: abridgement summarizes the narrative event (suspecting maid of placing letter) conveyed by the spoken utterances in the original text (``A letter has been placed... she said to her maid.'') \\
\hline
``If you will allow me, I have the key,'' said Grey. Then they both entered the house, and Vavasor followed his host up-stairs. & Mr. Grey unlocked the door of his house, and Vavasor followed him upstairs. & Commonsense inference: abridgement involves knowledge that doors are unlocked by keys, which is not explicit in the original text\\
\hline
George Osborne was somehow there already (sadly "putting out" Amelia, who was writing to her twelve dearest friends at Chiswick Mall), and Rebecca was employed upon her yesterday's work. & George Osborne was there already, and Rebecca was knitting her purse. & Narrative inference: ``knitting her purse'' in the abridgement is the event referenced by ``yesterday's work'' in the original, and resolving this requires knowledge of the previous text in the chapter\\
\hline
But Kate preferred the other subject, and so, I think, did Mrs. Greenow herself. & But Kate preferred the subject of the Captain, and so, I think, did Mrs. Greenow herself. & Elaboration: abridgement specifies ``Captain'' is the ``other subject'' implied in the original\\
\hline
\end{tabularx}
\caption{Examples of rows where alignment involves a language inference challenge}
\label{harder_examples}
\end{table*}

\subsection{Extended lexical category analysis}\label{extended_lexical_analysis}

Section \ref{lexical_categories} summarized the frequency of lexical categories for removed and added words in the \textsc{AbLit} test set, relative to these frequencies among all words in the original and abridged texts. Table \ref{removed_and_added_words_categories} additionally displays these percentages for all part-of-speech tags within the function and content word classes, along with examples of common words associated with each tag. We used the spacy library to perform part-of-speech tagging: \url{https://spacy.io/usage/linguistic-features}. 

\begin{table*}[h!]
\centering
\begin{tabularx}{\textwidth}{p{0.15cm} p{2.2cm} >{\columncolor{red!20}}p{1cm} >{\columncolor{red!20}}p{1.07cm} | >{\columncolor{green!20}}p{1cm} >{\columncolor{green!20}}p{1cm} X}
\textbf{Category} & & \textbf{\%$O$} & \textbf{\%$O_{rmv}$} & \textbf{\%$A$} & \textbf{\%$A_{add}$} & \textbf{Examples of Common Words}\\
\hline
\multicolumn{2}{l}{\textbf{Function words}} & \textbf{$\Sigma$=58.2} & \textbf{$\Sigma$=57.9} & \textbf{$\Sigma$=58.1} & \textbf{$\Sigma$=53.9}\\
& Punctuation & 14.0 & 12.9 & 15.7 & 23.3 & \texttt{,}\phantom{...}
\texttt{"}\phantom{...}
\texttt{.}\phantom{...}
\texttt{-{}-}\phantom{...}
\texttt{;}\phantom{...}
\texttt{-}\phantom{...}
\texttt{?}\phantom{...}
\texttt{!}\\
 & Pronoun & 11.0 & 10.2 & 11.5 & 8.7 & \texttt{i}\phantom{...}
\texttt{he}\phantom{...}
\texttt{it}\phantom{...}
\texttt{his}\phantom{...}
\texttt{her}\\
& Adposition & 10.2 & 11.5 & 9.0 & 7.0 & \texttt{of}\phantom{...}
\texttt{in}\phantom{...}
\texttt{to}\phantom{...}
\texttt{with}\phantom{...}
\texttt{for}\\
& Determiner & 7.8 & 8.3 & 7.1 & 3.9 & \texttt{the}\phantom{...}
\texttt{a}\phantom{...}
\texttt{an}\phantom{...}
\texttt{no}\phantom{...}
\texttt{all}\\
& Aux. Verb & 6.4 & 6.0 & 6.5 & 3.8 & \texttt{was}\phantom{...}
\texttt{had}\phantom{...}
\texttt{be}\phantom{...}
\texttt{is}\phantom{...}
\texttt{been}\\
& Coord. Conj. & 3.7 & 4.0 & 3.4 & 2.1 & \texttt{and}\phantom{...}
\texttt{but}\phantom{...}
\texttt{or}\phantom{...}
\texttt{nor}\phantom{...}
\texttt{both}\\
& Particle & 2.7 & 2.6 & 2.7 & 2.2 & \texttt{to}\phantom{...}
\texttt{not}\phantom{...}
\texttt{'s}\phantom{...}
\texttt{n't}\phantom{...}
\texttt{'}\\
& Subord. Conj. & 2.3 & 2.5 & 2.2 & 2.8 & \texttt{that}\phantom{...}
\texttt{as}\phantom{...}
\texttt{if}\phantom{...}
\texttt{when}\phantom{...}
\texttt{upon}\\
                    
\hline
\multicolumn{2}{l}{\textbf{Content words}} & \textbf{$\Sigma$=41.8} & \textbf{$\Sigma$=42.1} & \textbf{$\Sigma$=41.9} & \textbf{$\Sigma$=46.1} &\\
& Noun & 14.5 & 15.4 & 13.7 & 14.0 & \texttt{time}\phantom{...}
\texttt{man}\phantom{...}
\texttt{day}\phantom{...}
\texttt{way}\phantom{...}
\texttt{hand}\\ 
& Verb & 10.4 & 10.3 & 11.1 & 17.1 & \texttt{said}\phantom{...}
\texttt{had}\phantom{...}
\texttt{know}\phantom{...}
\texttt{do}\phantom{...}
\texttt{have}\\
& Adjective & 6.6 & 6.9 & 6.2 & 5.3 & \texttt{little}\phantom{...}
\texttt{own}\phantom{...}
\texttt{other}\phantom{...}
\texttt{such}\phantom{...}
\texttt{good}\\
& Adverb & 5.0 & 5.3 & 4.7 & 5.3 & \texttt{so}\phantom{...}
\texttt{very}\phantom{...}
\texttt{as}\phantom{...}
\texttt{now}\phantom{...}
\texttt{then}\\
& Proper Noun & 4.4 & 3.3 & 5.2 & 3.4 & \texttt{mr.}\phantom{...}
\texttt{mrs.}\phantom{...}
\texttt{sir}\phantom{...}
\texttt{miss}\phantom{...}
\texttt{lady}\\
& Other & 1.0 & 0.9 & 1.0 & 1.1 & 
\texttt{one}\phantom{...}
\texttt{two}\phantom{...}
\texttt{oh}\phantom{...}
\texttt{no}\phantom{...}
\texttt{yes}\\
\hline
\end{tabularx}
\caption{Distribution of part-of-speech categories for the set of all removed words $O_{rmv}$ and all added words $A_{add}$ in the \textsc{AbLit} test chapters. These numbers are respectively compared alongside those for the total set of all original words $O$ and all abridged words $A$. (Aux.=Auxiliary, Coord.=Coordinating, Conj.=Conjunction, Subord.=Subordinate)}
\label{removed_and_added_words_categories}
\end{table*}

\subsection{Comment about passage size variation}\label{model_inputs_comment}

Because the method for converting rows into passages of a consistent length (i.e. sentences, paragraphs, chunks) relies on string matching, the boundaries of the abridged passage may be off by one or a few words, which occurs less frequently as the size of the passages increase. This tends to occur when a word at the end of the original passage is replaced by a synonym in the abridged passage. However, a manual review of our assessment set revealed that only 0.4\% of sentences in the original text yielded abridgements with imprecise boundaries, and no paragraphs (and consequently no chunks) had this issue.

\subsection{Details about binary prediction model}\label{prediction_model_details}

For all passage sizes, we initialized models with the \textsc{RoBERTa-base} weights using the HuggingFace Transformers implementation: \url{https://huggingface.co/docs/transformers/v4.16.2/en/model_doc/roberta#transformers.RobertaModel}. \textsc{RoBERTa-base} consists of 125M parameters (\url{https://huggingface.co/roberta-base}). The maximum sequence length allowed by this model is 512, so we truncated all input tokens beyond this limit. We fine-tuned each model for 5 epochs, saving model weights after each epoch of training, and selected the model with the highest F1 score on the development set to apply to our test set. We used the AdamW optimizer \cite{loshchilov2017decoupled} and a batch size of 16. It took $\approx$2 hours to train each model on a g4dn.2xlarge AWS instance. During evaluation, any input tokens beyond the model length limit were assigned the default label of \texttt{preserved}. The result for each model reported in Table \ref{prediction_results} is based on a single run of the training procedure.

\subsection{Details about generation models}\label{generation_models_details}

Both \textsc{TunedT5} and \textsc{TunedBART} were fine-tuned using the HuggingFace transformers library, in particular this script: \url{http://github.com/huggingface/transformers/blob/master/examples/pytorch/summarization/run\_summarization.py}. \textsc{TunedT5} was initialized from \textsc{T5-base} \cite{raffel2020exploring}, which consists of $\approx$220M parameters (\url{https://huggingface.co/t5-base}). For this model, we prepended the prefix ``summarize: '' to the target (i.e. the abridged passage), consistent with how \textsc{T5-base} was trained to perform summarization. \textsc{TunedBART} was initialized from \textsc{BART-base} \cite{lewis-etal-2020-bart}, which consists of 140M parameters (\url{https://huggingface.co/facebook/bart-base}). For both \textsc{TunedT5} and \textsc{TunedBART}, we used a maximum length of 1024 for both the source (original passage) and target (abridged passage), and truncated all tokens beyond this limit. We evaluated each model on the development set after each epoch and concluded training when cross-entropy loss stopped decreasing, thus saving the model weights with the optimal loss. We used a batch size of 4. For all other hyperparameters we used the default values set by this script, which specifies AdamW for optimization. It took $\approx$3 hours to train each model on a g4dn.4xlarge AWS instance. The result for each model reported in Table \ref{generation_results} is based on a single run of the training procedure.

\subsection{Details about evaluation metrics}\label{evaluation_metrics_detail}

\paragraph{Preservation:} The formal definition of the preservation metric $Prsv$ is as follows. If $o_{prsv}(a_{pred})$ are the words in the original that are preserved in the predicted abridgement, and $o_{prsv}(a_{ref})$ are the words in the original that are preserved in the reference abridgement, then we consider the number of correctly preserved words: ${Correct\_Prsv = |o_{prsv}(a_{pred}) \cap o_{prsv}(a_{ref})|}$. The precision of this measure ${Prsv_p = \frac{Correct\_Prsv}{o_{prsv}(a_{pred})}}$ is the proportion of correctly preserved words among all preserved words in the predicted abridgement. The recall ${Prsv_r = \frac{Correct\_Prsv}{o_{prsv}(a_{ref})}}$ is the proportion of correctly preserved words among all preserved words in the reference abridgement. $Prsv$ is the F1 of these precision and recall measures: $Prsv = 2\frac{Prsv_p \cdot Prsv_r}{Prsv_p + Prsv_r}$.

\paragraph{Removal:} The formal definition of the removal metric is as follows. If $o_{rmv}(a_{pred})$ are the words in the original that are removed in the predicted abridgement, and $o_{rmv}(a_{ref})$ are the words in the original that are removed in the reference abridgement, then we consider the number of correctly removed words: ${Correct\_Rmv = |o_{rmv}(a_{pred}) \cap o_{rmv}(a_{ref})|}$. The precision of this measure ${Rmv_p = \frac{Correct\_Rmv}{o_{rmv}(a_{pred})}}$ is the proportion of correctly removed words among all removed words for the predicted abridgment. The recall ${Rmv_r = \frac{Correct\_Rmv}{o_{rmv}(a_{ref})}}$ is the proportion of correctly removed words among all removed words for the reference abridgement. $Rmv$ is the F1 of these precision and recall measures: $Rmv = 2\frac{Rmv_p \cdot Rmv_r}{Rmv_p + Rmv_r}$.

\paragraph{Addition:} The formal definition of the addition metric is as follows. If $a_{add}(a_{pred})$ are the words in the predicted abridgement that do not appear in the original, and $a_{add}(a_{ref})$ are the words in the reference abridgement that do not appear in the original, then we consider the number of correctly added words: ${Correct\_Add = |a_{add}(a_{pred}) \cap a_{add}(a_{ref})|}$. The precision of this measure ${Add_p = \frac{Correct\_Add}{a_{add}(a_{pred})}}$ is the proportion of correctly added words among all added words in the predicted abridgement. The recall ${Add_r =\frac{Correct\_Add}{a_{add}(a_{ref})}}$ is the proportion of correctly added words among all added words in the reference abridgement. $Add$ is the F1 of these measures: $Add = 2\frac{Add_p \cdot Add_r}{Add_p + Add_r}$. 

\subsection{Comment about addition scores}\label{comment_add_score}

Regarding the above-zero scores of the extractive methods on the $Add$ metric, there are two reasons for this. One reason is that the prediction model uses sub-tokens while the $Add$ metric analyzes whitespace-separated words. Consequently, one sub-token may be predicted as \texttt{preserved} while others within the same word are predicted as \texttt{removed}. Isolated from these other sub-tokens, the \texttt{preserved} sub-token will be recognized as a new added word in the abridgement. The other reason is that a single word in the original may be split by the tokenizer into two words in the abridgement, or vice-versa. For example, we observed that ``Mr.'' gets split into two tokens (``Mr'', `.') in some contexts and is treated as one token (``Mr.'') in others. If the original text represents this item as two tokens and both the extracted and reference abridgement represent it as a single token, then this single token will be counted as an added word in the extracted abridgement.

\subsection{Examples of produced abridgements}\label{produced_abridgement_examples}

Tables \ref{abridgement_examples1} and \ref{abridgement_examples2} below show excerpts of the abridgements produced by the \textsc{ExtSent} and \textsc{TunedBART} models, alongside the original chapter text and human-authored reference abridgement. The sentences in each excerpt are lined up to better visualize their differences.

\begin{table*}[h!]
\centering
\begin{tabularx}{\textwidth}{>{\columncolor{gray!10}}X | >{\columncolor{green!10}}X | >{\columncolor{orange!10}}X | >{\columncolor{violet!10}}X}
\textbf{Original} & \textbf{Reference} & \textbf{\textsc{ExtSents}} & \textbf{\textsc{TunedBART}}\\
\hline
Seven days glided away, every one marking its course by the henceforth rapid alteration of Edgar Linton's state. & In the next seven days Edgar Linton's state grew rapidly worse. &  & Seven days glided away, every one marking its course by the rapid alteration of Edgar Linton's state.\\
The havoc that months had previously wrought was now emulated by the inroads of hours. & & & The havoc that months had previously wrought was now emulated by the inroads of hours.\\
Catherine we would fain have deluded yet; & Catherine could no longer be deluded: & & Catherine we would fain have deluded yet;\\
but her own quick spirit refused to delude her: & & but her own quick spirit refused to delude her: &  but her own quick spirit refused to delude her:\\
it divined in secret, and brooded on the dreadful probability, gradually ripening into certainty. &  she brooded on the dreadful probability of her father's death, gradually ripening into certainty. & & it brooded on the dreadful probability, gradually ripening into certainty.\\
She had not the heart to mention her ride, when Thursday came round; & She had not the heart to mention her ride when Thursday came round. & She had not the heart to mention her ride, when Thursday came round; & She had not the heart to mention her ride, when Thursday came round;\\
I mentioned it for her, and obtained permission to order her out of doors: & I obtained permission to send her out of doors: & I mentioned it for her, and obtained permission to order her out of doors: & I ordered her out of doors:\\
for the library, where her father stopped a short time daily-the brief period he could bear to sit up-and his chamber, had become her whole world. & for her father's chamber had become her whole world. & &  for the library, where her father stopped daily - the brief period he could bear to sit up - and his chamber, had become her whole world.\\
She grudged each moment that did not find her bending over his pillow, or seated by his side. & She grudged each moment that she did not spend bending over his pillow, or seated by his side. & & She grudged each moment that did not find her bending over his pillow, or seated by his side.\\
Her countenance grew wan with watching and sorrow, and my master gladly dismissed her to what he flattered himself would be a happy change of scene and society; & She grew pale with watching, and my master gladly dismissed her to what he thought would be a happy change of scene; & Her countenance grew wan with watching and sorrow, and my master gladly dismissed her to what he flattered himself would be a happy change of scene and society; & Her countenance grew wan with watching and sorrow, and my master gladly dismissed her to what he flattered himself would be a happy change of scene and society;\\
drawing comfort from the hope that she would not now be left entirely alone after his death. & drawing comfort from the hope that she would not now be left entirely alone after his death. & drawing comfort from the hope that she would not now be left entirely alone after his death. & drawing comfort from the hope that she would not now be left entirely alone after his death.\\
\hline
\end{tabularx}
\caption{Abridgements for an excerpt of Wuthering Heights, Chapter 27}
\label{abridgement_examples1}
\end{table*}

\begin{table*}[h!]
\centering
\begin{tabularx}{\textwidth}{>{\columncolor{gray!10}}X | >{\columncolor{green!10}}X | >{\columncolor{orange!10}}X | >{\columncolor{violet!10}}X}
\textbf{Original} & \textbf{Reference} & \textbf{\textsc{ExtSents}} & \textbf{\textsc{TunedBART}}\\
\hline
It happened that when I came home from Deal I found a note from Caddy Jellyby (as we always continued to call her), informing me that her health, which had been for some time very delicate, was worse and that she would be more glad than she could tell me if I would go to see her. & When I came home from Deal I found a note from Caddy, informing me that her health, which had been for some time very delicate, was worse and that she would be very glad if I would go to see her. & It happened that when I came home from Deal I found a note from Caddy Jellyby (as we always continued to call her), informing me that her health, which had been for some time very delicate, was worse and that she would be more glad than she could tell me if I would go to see her. & It happened that when I came home from Deal I found a note from Caddy Jellyby informing me that her health, which had been for some time very delicate, was worse and that she would be more glad than she could tell me if I would go to see her.\\

It was a note of a few lines, written from the couch on which she lay and enclosed to me in another from her husband, in which he seconded her entreaty with much solicitude. & It was a short note, written from her bed. & & It was a note of a few lines, written from the couch on which she lay and enclosed to me in another from her husband, in which he seconded her entreaty with much solicitude. \\

Caddy was now the mother, and I the godmother, of such a poor little baby--such a tiny old-faced mite, with a countenance that seemed to be scarcely anything but cap-border, and a little lean, long-fingered hand, always clenched under its chin. & Caddy was now the mother, and I the godmother, of such a poor little baby - such a tiny old-faced mite, with a little lean, long-fingered hand always clenched under its chin. & Caddy was now the mother, and I the godmother, of such a poor little baby--such a tiny old-faced mite, with a countenance that seemed to be scarcely anything but cap-border, and a little lean, long-fingered hand, always clenched under its chin. & Caddy was now the mother, and I the godmother, of such a poor little baby - such a tiny old-faced mite, with a countenance that seemed to be scarcely anything but cap-border, and a little lean, long-fingered hand, always clenched under its chin.\\

It would lie in this attitude all day, with its bright specks of eyes open, wondering (as I used to imagine) how it came to be so small and weak. & It would lie in this attitude all day, with its bright specks of eyes open, wondering (I used to imagine) how it came to be so small and weak. & It would lie in this attitude all day, with its bright specks of eyes open, wondering (as I used to imagine) how it came to be so small and weak. & It would lie in this attitude all day, with its bright specks of eyes open, wondering how it came to be so small and weak.\\

Whenever it was moved it cried, but at all other times it was so patient that the sole desire of its life appeared to be to lie quiet and think. & Whenever it was moved it cried, but at all other times it lay quiet. &  Whenever it was moved it cried, but at all other times it was so patient that the sole desire of its life appeared to be to lie quiet and think. & Whenever it was moved it cried, but at all other times it was so patient that the sole desire of its life appeared to be to lie quiet and think.\\


\hline
\end{tabularx}
\caption{Abridgements for an excerpt of Bleak House, Chapter 50}
\label{abridgement_examples2}
\end{table*}

\end{document}